\documentclass[letterpaper, 10 pt, conference]{ieeeconf}  

\IEEEoverridecommandlockouts                              
\UseRawInputEncoding
\usepackage{multirow}
\usepackage{subfig}
\usepackage[linesnumbered,lined,boxed,commentsnumbered]{algorithm2e}
\usepackage{caption}

\usepackage{cite}
\usepackage{amsmath,amssymb,amsfonts}
\usepackage{algorithmic}
\usepackage{graphicx}
\usepackage{textcomp}
\usepackage{xcolor}

\usepackage{graphicx}
\usepackage{booktabs}
\usepackage{pgfplots}
\usepackage{amsmath}
\usepackage{float}
\usepackage{multirow}

\title{\LARGE \bf
Helping Language Models Learn More: Multi-dimensional Task Prompt for Few-shot Tuning*
}

	\author{Jinta Weng, Jiarui Zhang, Yue Hu$^{*1}$, Daidong Fa$^{2}$, Xiaofeng Xu$^{3}$ and Heyan Huang$^{*4}$
\thanks{*This work is supported by the National Natural Science Foundation of China (Grant No.U21B2009), Guizhou Province Education Department Project under Grant Qianjiaohe KY[2022]091, Qiannan Science and Technology Planning Project under Grant (No.[2018]19), and is also supported by the Strategic Priority Research Program of Chinese Academy of Science, Grant No.XDC02030400.}
\thanks{$^{1}$Jinta Weng, Jiarui Zhang, Yue Hu(corresponding author) is with the School of Cyber Security, University of Chinese Academy of Sciences, China
        {\tt\small wengjinta@iie.ac.cn, zhangjiarui@iie.ac.cn, huyue@iie.ac.cn}}%
\thanks{$^{2}$ Daidong Fa with the School of Computer and Information, Qiannan Normal University for Nationalities, Duyun, China
        {\tt\small FDD19870115@yeah.net}}%
\thanks{$^{3}$ Xiaofeng Xu with the School of Computer Science and Cyber Engineering, Guangzhou University, Guangzhou, China
   	{\tt\small xfxu.cs@outlook.com}}%
\thanks{$^{4}$ HeYan Huang, corresponding author, with the School of Computer Science and Technology, the Southeast Academy of Information Technology, Beijing Institute of Technology, China {\tt\small hhy63@bit.edu.cn}}
}

\begin{document}

\maketitle
\thispagestyle{empty}
\pagestyle{empty}

\begin{abstract}
Large language models (LLMs) can be used as accessible and intelligent chatbots by constructing natural language queries and directly inputting the prompt into the large language model. However, different prompt' constructions often lead to uncertainty in the answers and thus make it hard to utilize the specific knowledge of LLMs (like ChatGPT). To alleviate this, we use an interpretable structure to explain the prompt learning principle in LLMs, which certificates that the effectiveness of language models is determined by position changes of the task's related tokens. Therefore, we propose MTPrompt, a multi-dimensional task prompt learning method consisting based on task-related object, summary, and task description information. By automatically building and searching for appropriate prompts, our proposed MTPrompt achieves the best results on few-shot samples setting and five different datasets. In addition, we demonstrate the effectiveness and stability of our method in different experimental settings and ablation experiments. In interaction with large language models, embedding more task-related information into prompts will make it easier to stimulate knowledge embedded in large language models.

\end{abstract}

	\section{Introduction}
\begin{figure}[htbp]
	\centerline{\includegraphics[scale=0.35]{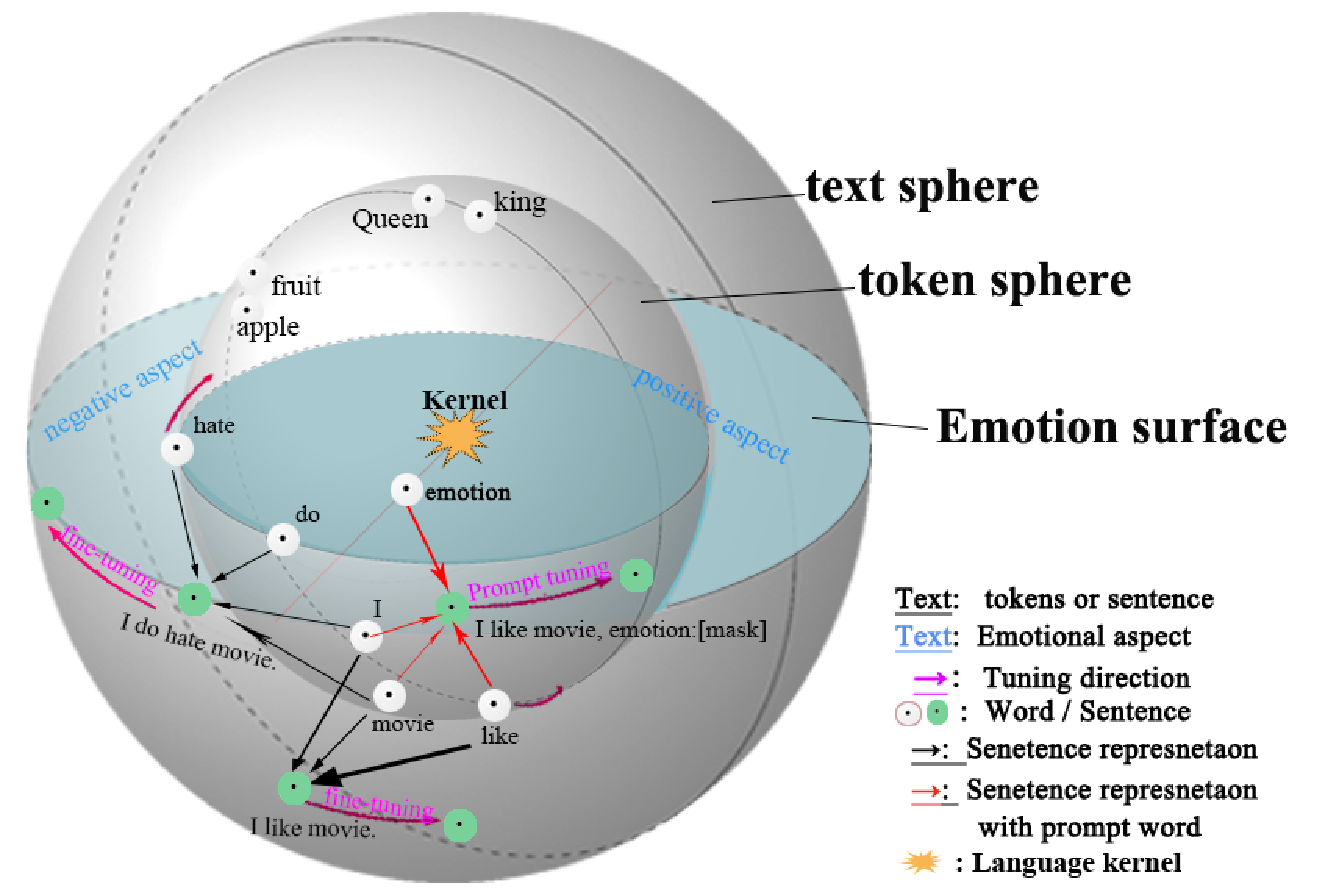}}
	\caption{The analytic structure of prompt-tuning and fine-tuning processes.}
	\label{figure1}
\end{figure}

Pre-trained language models (PLMs) have widely changed the research paradigm and applications(like ChatGPT and GPT4) of natural language processing in recent years \cite{2018bert}. At present, most of these models are composed of two processes: pre-training model and fine-tuning process. PLMs use its huge parameters to record or learn the linguistic relation within large open-resource corpus in pre-training process. Then, with limited fine-tuning parameters and more domain-related datasets, PLMs can show better task-level adaption on downstream application.

Ideally, linguistic knowledge and semantic relations are re-parameterized and motivated in the PLMs. Thus, through the proper design of external prompt, the PLMs' knowledge can be extracted or activated. Motivated by this, different model designs and promoting methods, like GPT-3's demonstration learning \cite{Brown2020Language} and cloze question templates \cite{jiang2020can}, have shown their powerful ability in few-shot learning tasks. 
%
In these prompt-based fine-tuning methods, the model generally consists of a manual template to generate an explicit or implicit prompting input, a mask token for label prediction, and the label mapping that maps different labels into specified words. 
\par
However, finding a suitable template is fundamental to utilizing the prompt in PLMs \cite{lester2021power,kavumba2022prompt}. Even minor change in the prompt form can make the result widely different \cite{qiu_pre-trained_2020}. Most researchers try to explore suitable prompts from the aspect of their representation and model construction. For example, Schick and Schütze propose the PET and IPET models from the viewpoint of multi-template learning and using the expanded training dataset, but the effectiveness is based on hidden information of large unlabeled datasets \cite{schick2021exploiting}. Chen et al.\cite{gao2020making} also design a violence searching method for selecting the suitable label mapping, which allows generating maximal hidden value of mask token on all inputs . Webson et al.\cite{webson2021} argues that most prompt-based fine-tuning methods are more like relaxed RegEx rules, which aim to fit the language model and original corpora. Moreover, the search space for finding a suitable template or label mapping explosively increases with the number of tokens or labels. Nevertheless, the effectiveness is mainly based on the experienced manual-defined template, which makes prompt design still a specialist-first and computational-ability-first thing due to its heuristic-discovery process. 

Unlike the textual prompt above, by encoding the prompt in a continuous vector in PLMs, continuous prompt (also called soft prompt) learning methods have effectively reduced the cost of finding suitable prompt \cite{liu2021gptunderstand,li-liang-2021-prefix,shin2020autoprompt,schick2021exploiting}. These continuous prompts can tune the prompt vector and learn the hidden relationships within the template and the original input, while it may lack portability and interpretation. Moreover, the suitable initialization of the continuous prompt is also essential to realize an ideal effect, which is more common in few-shot learning tasks \cite{webson2021}. 

In general, although some prompt learning methods have achieved significant effects, the efficiency and reliability of prompt learning are still considered issues. It's not easy to quickly design an effective and usable prompt.



\par
To get close to this problem, we propose a sphere model to reveal the learning mechanism of pre-trained language models (PLMs). Based on the proposed structure, we explain the effectiveness of the prompt tuning process by comparing the fine-tuning process. We reveal that adding appropriate task-related tokens could facilitate sentence representation in a more appropriate position. Therefore, we proposed the MTPrompt model, a \underline{M}ulti-dimensional \underline{T}ask-driven \underline{P}rompt learning model generated from different types of task descriptions. We aim to design more instructive prompts from the degree of meta-description of the task and help the model utilize more task-oriented information from PLMs. 
The main contributions of the proposed paper are as follows:
\begin{itemize}
	\item We reveal the learning and prompting mechanisms of natural language models by the interpretative modeling method. 
	\item A multi-dimensional task prompt used to stimulate the information activation of language models is proposed. 
	\item Our proposed method can achieve state-of-the-art results in various few-shot learning tasks.
\end{itemize}

\section{Theory Foundation}
In this section, we first introduce the definition of fine-tuning and prompt-based tuning. Then we use the analytic structure method to clarify the theoretical basis of the proposed MTPrompt. 
\subsection{Fine-tuning Process}
Fine-tuning is the most basic model-tuning method for natural language processing applications. By initializing the language model with pre-trained parameters (acquired from an open-source trained model), the fine-tuning process tunes these parameters on task-related data with a gradient descent training process.
Therefore, large language models can quickly adapt to downstream tasks, and fine-tuning has also become a standard trick for deep learning and neural networks. It has been shown that using a pre-trained model on a large dataset and fine-tuning on task-related data can often get better results than training directly on your data without the pre-trained model because the parameters of the pre-trained model are in a better position for the fine-tuning process.

\subsection{Prompt-based fine-tuning}
With the exponential growth of the pre-trained language model, large language models (such as Chat-GPT and GPT3), which can store human knowledge and adjust to changes in the natural language query (Prompt), could complete some tasks and realize human interaction based on parameter emergence ability. Therefore, a tuning method called prompt-based fine-tuning (prompt tuning) has been proposed. 

The prompt-based fine-tuning consists of a human-defined template and label verbalizer. For the text emotion classification task, we may use a template like `input. Emotion:[mask position]' to transform the input 'I like movie' into `I like movie. Emotion:[mask position]'. The [mask position] is used to generate specific words like `positive' or `negative'. If the prediction word of mask position is `positive', we would directly classify the input as positive emotion while the word `negative' is for negative emotion. The prediction word could be changed based on the task's mapping labels, and we call this label mapping process a label verbalizer. The pre-trained model only needs to predict the mask position in specific words with the help of a prompt template and a verbalizer.

\begin{figure}[htbp]
	\centerline{\includegraphics[scale=0.3]{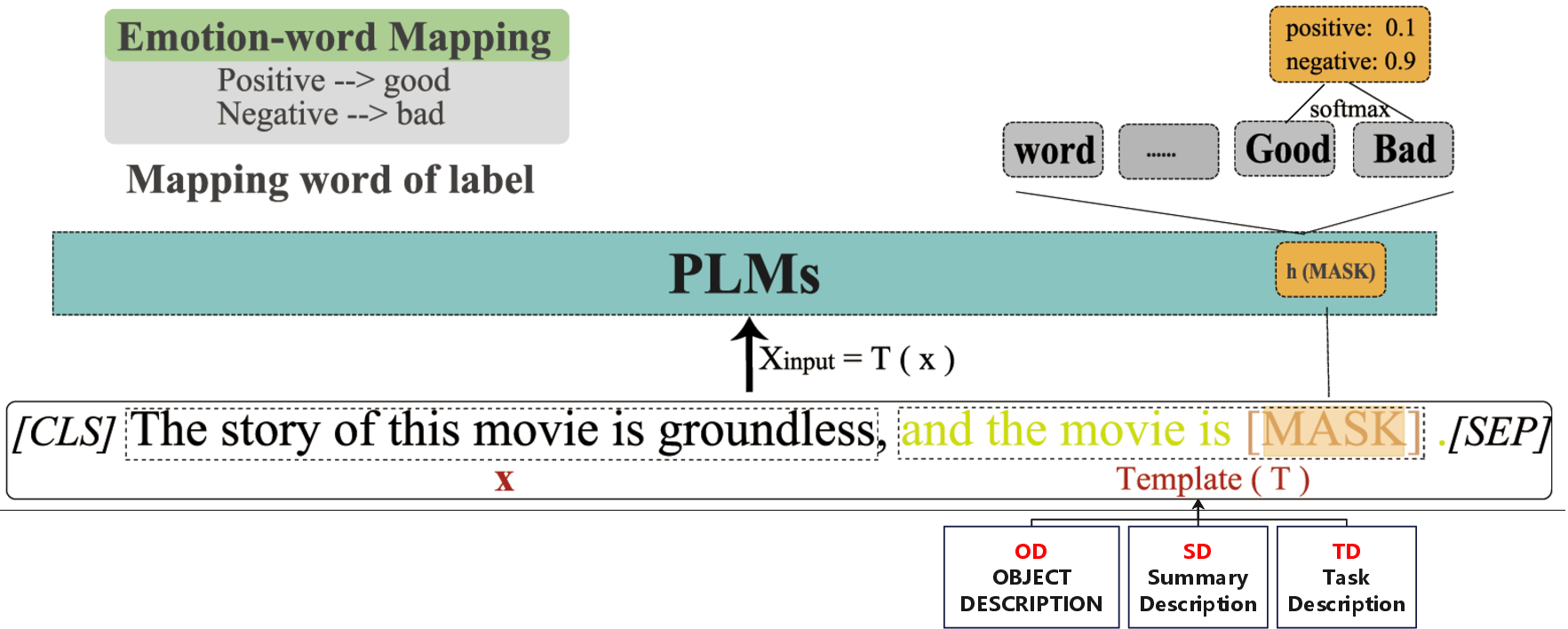}}
	\caption{The prompt tuning procession with MTPrompt.}
	\label{figure2}
\end{figure}
\subsection{Prompt Mechanism and Limitation}
To reveal the different and the association of the fine-tuning and the prompt-based fine-tuning method in PLMs. As figure \ref{figure1} depicted, we construct the analytic structure between the fine-tuning and the prompt-tuning process. 

We first classify the pre-trained language model into three categories, the language kernel, token sphere, and text sphere. The language kernel is the ideal language origin with language syntax. All the words are listed in the token sphere with different positions, and the distance between them reveals their semantic similarity(Like apple and fruit). The text sphere with maximum space is used to represent different sentences, and the positions of these sentences are calculated by the words in the token sphere.
The existence of the concept of `position' is a task-specific perspective. For example, in the emotional classification task, we define an emotional classification surface to classify these words and sentences into suitable positions. For example, the projection of the word `like' on the emotional surface should be closer to the positive aspect. 

In the fine-tuning process, the sentence `I like movie' and the word `like' would fine-tune to the positive aspect, while the sentence `I hate movie' and the word `hate' would fine-tune to the negative aspect.

Different from the fine-tuning process, the input would first add a prompt template with tokens `emotion:', then the position of the sentence `I like to move' is influenced by the adding token `emotion', which brings a new initialization position closer to the emotional surface. Therefore, prompt-based fine-tuning needs less training corpus and less training cost with a more suitable initialization position, as the information of adding prompt template and tokens can influence the finnal prediction. 

What is more, the effectiveness of prompting learning method have shown PLM is a parameterized knowledge base of training corpus. The prompt-based fine-tuning process not only optimizing huge parameters to adapt in downstream task, but also finding a best parameterized path to access similar knowledge. Furthermore, could we optimize the prompt-tuning process through more task-related tokens  and task-related information?

\section{MTPrompt: Multi-dimensional Task Prompt}

Motivated by above-mentioned discovery, we put forward a new prompt learning, Multi-dimensional Task Prompt(MTPrompt). In detail, the MTPrompt is composed of three different types of description: object description (OD), summary description (SD), and task description (TD). 
These descriptions are generated from the metadata of current task, such as label information, task goal, and other metadata information, which are easy to extracted from opensource knowledge.

\begin{table}[H]
	\scalebox{0.98}{
		\centering
		\begin{tabular}{p{7cm}}
			\toprule
			\textbf{Task name (labels)}\\  - Prompt Type: Description \\                   
			\midrule
			\textbf{Example: SST-2 Task} (positive / negative) \\ \textbf{- OD}: A movie review \\ \textbf{- SD}: Talking about its director, actor, performance, character skill, and story. \\ \textbf{- TD}: The emotion of this review was [MASK]  \\
			\midrule
		\end{tabular}
	}
	\caption{Our Meta-prompt combining by object description (OD), summary description (SD), and task description (TD) with a `mask' token.}
	\label{tab:tabel1}	
\end{table}

We list a MTPrompt of SST-2(The Stanford Sentiment Treebank) emotional classification tasks in Tab. \ref{tab:tabel1}. 

As following equation depicted, after select suitable types of MTPrompt, the original input $ x^{i} $ is transferred to:
\begin{equation}
T_{}(x^{i})= x^{i}.[t_{Od}][t_{Sd}][t_{Td}][MASK]
\label{eq:meta}
\end{equation}
where T is the template for the input x, i is the index of the instance, and t are the different types of word tokens.
\par

Since PLMs may see similar tasks and corpus in the pre-training process, we could use more metadata description about current task to help the model recall its knowledge memory quickly, instead of merely using a cloze question or demonstration example like GPT-3. Besides, the metadata description could help the PLMs fine-tune in more narrow and similar semantic space, for which the metadata description template introduces more task-related information.

%
%

\begin{table}[htb]
	\centering
	\scalebox{1}{\begin{tabular}{l}
			\toprule
			\textbf{Task name (labels)} \\-[OD]-[SD]-[TD])\\                   
			\midrule
			\textbf{SST-2 } (positive / negative) \\ - A movie review \\ - talking about its director, actor, performance, character skill, and story,\\ - the emotion of this review was [MASK]  \\
			\midrule
			\textbf{SST-5} (very positive / positive / neutral / negative / very negative)  \\  - A movie review \\ - talking about its director, actor, performance, character skill, and story,\\ - the emotion of this review was [MASK]  \\
			\midrule
			\textbf{TREC}  (abbreviation / entity / description / human / location / numeric)  \\  	
			- A English question.  \\
			- about huaman, description, location numeric entity , and abbreviations. \\
			- The question type is [mask].  \\
			\midrule
			\textbf{SNLI } (entailment / neutral / contradiction)  \\ 
			- $[sent_{1}.]$  A Stanford Natural Language Inference sentence pairs \\	
			- labeled as entailment , contradiction , and neutral. \\	
			- whether the context contains the answer to the question?[mask]$[sent_2.]$ \\
			\midrule
			\textbf{QNLI} (entailment / contradiction)\\		
			- $[sent_{1}.]$  A Stanford Question Answering sentence pairs \\	
			- labeled as entailment and contradiction. \\	
			- whether the context contains the answer to the question? [mask]$[sent_2.]$
			\\
			\bottomrule
	\end{tabular}}
	\caption{Our MTPrompt combining object description (OD), summary description (SD), and task description (TD) with a mask token.}
	\label{Tab:C1}
\end{table}
\subsection{Select Semantic Prompt}
Since minor change of prompt tokens may result in the uncertainty result, thus selecting suitable template $  T$(prompt) is essential. We thus introduce the prompt candidate to store different types of MTPrompts and put forward a searching strategy for the prompt selection. We provide manual and automatic selection, allowing users to select a specific type of prompt or to traverse supported prompts. The formula of automatic selection process is as follows:
\begin{equation}
topk(\sum_{i}^{D} p\left ( y| T_n(x_{i}), mask \right ))
\end{equation} 
where D is the size of task's training sample, n is the index of different templates, y is the label?s mapping token corresponding to the label of X.

In the manual setting, we have listed some ideal prompt options for experimental datasets in the table \ref{Tab:C1} by means of automatic searching. As the suitable prompt is selected, a transition template of current prompt is used to transform the original input $ x^{i} $ to prompting $ \overline{x}^{i} $:
\begin{equation}
\overline{x}^{i} = T_{select}(x^{i}) = x^{i}.[t_{1}][t_{2}][t_{..}][t_{N}][MASK].
\end{equation}
where the \textit{[mask]} is the specif token that would be subsequently used to generate the word distribution over PLM's vocabulary, and  $T_{select}$ is the selected template.
\par

\subsection{Model formulation}
Given the transition input $\overline{x}^{i}$, the target is predicting the pre-definded label-mapping tokens on \textit{[mask]} position.

As the Fig. \ref{figure2} depicted, only the output of \textit{[mask]} token $ h_{[mask]} $ would be used to create a token-level distribution over PLM's vocabulary by introducing a learnable linear projector W:
\begin{equation}
\delta  = h_{[mask]} \otimes W_{proj} =   PLM(\overline{x}^{i}_{mask} ) \otimes W_{proj} 
\end{equation}
where the size of linear projector W is $R^{{|h}\times{len(V)|}}$, and $h_{v_{t}}$ is the value of PLM token distribution in token $v_{t}$.

Subsequently, some tokens in PLM's vocabulary $  V $ are chosen to represent each label and formed in a pre-defined label mapping.
\begin{equation}	
F(y): y_{t} \rightarrow v_{t}, y_{t} \in {Y}, v_{t} \in {V}
\end{equation}
, where $ t $ is the label index of label set $ Y $, $y_{t}$ is the specific label, and $ v $ represents the token in Vocabulary $ V $. For example, in emotion classification task, we could use the PLMs token ``good'' to stand for the positive label, while using ``bad'' for negative label.
\par

Since each label has been represented by some specific tokens, only the distributions of these tokens $v_{t}$ would be retained and used to calculate the final prediction as following equation:

\begin{equation}
\begin{split}
p(y^{i}|\overline{x}^{i}) & \Rightarrow p(F(y^{i})|	\delta ) = p(v^t | 	h_{[mask]} \otimes W_{proj}) \\
& = \ln \dfrac{\exp(h_{[mask]} \cdot w_{t})}{\begin{matrix} \sum_{k=1}^{|Y|}\exp(h_{[mask]} \cdot w_{k}) \end{matrix}}
\end{split}
\end{equation}
, where $|Y|$ is the label number of current task, $v_{k}$ is used to represent the token?s logit value of the k-th token in soft label mapping. 

The target of task classification thus is transferred into the prediction of tokens in label mappings and the final loss is formulated as:
\begin{equation}
L=-\dfrac{1}{N}\sum_i \sum_{t=1}^{|Y|} \log{p(y^{it}|x^{it})}
\end{equation}
, where $i$ is the index of the training pair $(x_{i},y_{i})$.


\begin{table*}[htb]
	\centering
	\scalebox{0.9}{
		\begin{tabular}{l|ll}
			\hline
			\textbf{Task} & \textbf{Label Mapping}                                                          & \textbf{Template}                             \\ \hline
			SST-2         & \{'0':'good','1':'bad'\}                                                        & *cls**sent\_0*\_It\_was*mask*.*sep+*          \\
			SST-5         & \{`contradiction':`No',`entailment':`Yes',`neutral':`Maybe'\}                   & *cls**sent\_0*\_This\_movie\_was*mask*.*sep+* \\
			TREC          & \{0:`Description',1:`Entity',2:`Expression',3:`Human',4:`Location',5:`Number'\} & *cls**mask*:*+sent\_0**sep+*                  \\
			QNLI          & \{`not\_entailment':`No',`entailment':`Yes'\}                                   & *cls**sent-\_0*?*mask*,*+sentl\_1**sep+*      \\
			SNLI          & \{`contradiction':`No',`entailment':`Yes',`neutral':`Maybe'\}                   & *cls**sent-\_0*?*mask*,*+sentl\_1**sep+*      \\ \hline
		\end{tabular}
	}
	\caption{The label mapping and template in the other baselines. |Y|: classes Number for classification tasks. L: average words in input sentence(s). In our few-shot experiments,  we also sample $ D_{train} $ and $D_{dev}$ of K × |Y| examples from the original training set.}
	\label{tab:tabel3}
\end{table*}

\begin{table*}[htbp]
	\centering
	\scalebox{1}{
		\centering
		\begin{tabular}{lccccc}
			\toprule
			\textbf{Baselines} \  &\textbf{ TREC (acc)} & \textbf{SNLI (acc)} &\textbf{ QNLI (acc)} & \textbf{SST-5 (acc)}& \textbf{ SST-2 (acc) } \\
			\midrule
			Majority         $ \star $                   & 18.8      & 33.8      & 49.5 & 50.9      & 23.1      \\
			prompt-based zero-shot learning$ \star $     & 32.0      & 49.5      & 50.8        & 35.0   & 83.6    \\
			GPT3-in-context-learning$ \star $            & 26.2(2.4) & 47.1(0.6) & 53.8(0.4)    & 30.6(0.9) & 84.8    \\
			fine-tuning$ \star $                         &88.8(2.1) & 48.4(4.8) & 60.2(6.5)  & 43.9(2.0) & 81.4(3.8) \\
			\midrule
			Prefix-tuning$  $ & 36.0(1.1) & 33.5(3.9) & 54.5(2.2) & 46.1(1.3)  & 88.1(2.3) \\
			P-tuning$  $ & 40.2(1.3) & 37.5(1.6) & 57.6(3.9) & 32.1(3.1)  & 90.1(1.2) \\
			
			LMBFF $ \star $ & 84.8(5.1) & 77.1(3.9) & 63.7(4.2) & 46.1(1.3)  & 92.1(1.1) \\
			CP-Tuning & -- & --& 69.22 & -- & 93.35\\
			UPT & 76.2 & 71.1 & -- & 45.1 & 90.8\\
			\midrule
			
			\textbf{Our Methods}      &   &  &    &     &  \\\midrule
			
			MT-prompt $\dagger$  & \textbf{85.0(2.0)} & \textbf{77.2 (1.4)} & 64.9 (3.3)  &  \textbf{50.4(1.3)} & 91.9  \textbf{ (1.5)}\\

			\bottomrule
			
		\end{tabular}
	}
	\caption{Main result of the MTPrompt. The results of all experiments are evaluated by selecting the mean and variance of accuracy on five different segmented training datasets and same testing dataset. Note that the results of existing baselines ($ \star $) are used from the \cite{gao2020making} to ensure fairness, and the results of prefix-tuning and prompt-tuning mode are reproduced by the same experimental setting.}
	\label{tab:tabel4}
\end{table*}

\section{Experiments}
In this section, we present the effectiveness of proposed model MTPrompt in different semantic understanding tasks with few-shot setting, for example, the sentiment classification, question classification, and natural language inference. We also try to explore and analysis the generalization and transferability  and Meta-prompt.
\subsection{Datasets}
We have selected five distinct datasets from the glue and sentiment benchmarks, which include tasks such as sentiment classification, question classification, and natural language inference. For the sentiment classification task, we use the two most representative datasets: SST-2 and SST-5. The SST datasets are used to predict the emotion of a movie review, with SST-2 labeling reviews as positive or negative and SST-5 labeling reviews with tags such as 'very positive', 'positive', 'neutral', 'negative', and 'very negative'. We have chosen the TREC-6(The Text REtrieval Conference Question Classification dataset) dataset for the question classification task, which aims to predict the six types of questions given an English question text. Finally, for the natural language inference task, we have utilized both the SNLI(Stanford Natural Language Inference) and QNLI(Qusetion-answering NLI)) datasets to evaluate the model's ability to determine whether the meaning of the next sentence can be inferred from the previous sentence.
\subsection{Experiment Setting}
Our pre-training language model is RoBERTa-large. The experiments were conducted on NVIDIA V100 32GB, although they could also be run on 1080ti with a small batch size. During the training process, we conducted various experiments with different batch sizes (bs=4,8,16) and learning rates (lr=1e-5,2e-5,5e-5). To ensure fair results, we used five different sub-sets with the same sampling size, as some studies have shown that even minor differences in few-shot datasets can lead to varied results. Additionally, to ensure fairness, we used the mean score and variance of the prediction result on different subsets instead of the highest score. We followed Gaos work to satisfy the few-shot learning in PLMs by selecting five different K-shot sub-datasets, where each sub-dataset consists of K=16 training pairs on each type of label\cite{gao2020making}. For instance, the SST-2 emotion classification task with two classes needed five different training sets with the size of 32 and a validation set with the same size of 32, while the testing size used the original size of the SST-2 task without any other data setup. These sub-datasets were trained and predicted individually, and their training process and prediction result were recorded with different batch sizes and learning rates. All training pairs in the same task used the same template transition in the proposed prompting strategies to construct the real input. Finally, we selected the best result of each sub-dataset using different hyperparameters and integrated the best result on the aspect of the sub-dataset. To evaluate the prediction of sub-datasets, we selected the average accuracy and variance of different sub-datasets as the evaluation criteria, as it shows the overall performance and variation of the current task.


\begin{table}[htb]
	\centering
	\scalebox{1.0}{
		\begin{tabular}{lccccc}
			
			\toprule
			\textbf{Combinations}   &\textbf{ Acc.} & \textbf{Variance} & \textbf{Median} \\
			\midrule
			OD+SD+TD w/0  & 46.1    & 1.3 & 46.4  \\\midrule
			TD  & 47.5   & 2.3 & 48.5  \\
			OD  & 48.2  & 2.5 & 48.4  \\
			SD         &48.3    & 2.1 & 47.1  \\\midrule
			SD+TD	  & 48.7$ \uparrow $  &  0.6$ \downarrow $   & 48.6$ \uparrow $  \\
			OD+TD        & 48.8$  \uparrow $     & 1.0$ \downarrow $ & 48.6$ \uparrow $  \\
			OD+SD       & 50.1  $ \uparrow $      & 1.4$ \downarrow $ &  49.8$ \uparrow $ \\\midrule	
			OD+SD+TD         & 47.3 $ \downarrow $   &  5.4  $ \uparrow $ & 50.0$  \uparrow $ \\\midrule
			
			\textbf{Average}   & 49.5 &1.7  &50.3$  \uparrow $\\
			\bottomrule
			
		\end{tabular}
	}
	\caption{The effectiveness of different types and combinations of task descriptions in Meta-prompt. The average result is generated from the joint inference of all combinations of Meta-prompt.}
	\label{tab:tabel5}
\end{table}

\begin{table*}[htb]
	\scalebox{1.0}{
		\begin{tabular}{lllllllllll}
			\hline
			\multirow{2}{*}{Task Name} & \multicolumn{10}{c}{Settings}                                                                                             \\ \cline{2-11} 
			& \textbf{bs=4} & \textbf{bs=8} & \textbf{bs=16} & \textbf{seed=13} & \textbf{seed=21}& \textbf{seed=42}& \textbf{seed=87} &\textbf{seed=100}& Average & LMBFF \\ \hline
			SST-2                       & 90.5 (2.2)            & \textbf{ 92.8 (0.2)}            & 91.7 (1.4)                         &  89.2 (0.0)     & \textbf{93.6 (0.0)}          &  92.5 (0.0)            & 91.6 (0.0)                       & 92.5 (0.0)   & 91.9 (1.5)  & 92.1(1.2)
			\\
			SST-5                      &   47.7 (2.6)         &  48.9 (1.9)    &\textbf{ 50.1 (1.1)}            &  50.3 (0.0)         & 48.1 (0.0)          &\textbf{ 51.4 (0.0)}             & 46.9 (0.0)                        &  50.5 (0.0) & 50.4 (1.3) &46.1(1.3)  \\ \hline
		\end{tabular}
	}
	\centering
	\caption{The comparison of different experimental settings. Note that the variance result is based on different seeds and therefore the variance of different seeds is null. }
	\label{tab:tabel6}
\end{table*}

\subsection{Baselines}

We evaluate our approach by comparing it to several existing methods and setting our baselines. The baselines we use include the Majority method, which selects the class with the highest frequency as the prediction, fine-tuning method \cite{Liu2019RoBERTaAR}, prompt-based zero-shot learning, GPT3-in-context-learning  \cite{Brown2020Language}, and the LMBFF model described in Gao's work \cite{gao2020making}. To ensure a fair comparison with the baselines, we adopt the same settings used in previous studies for each prompt learning method. The specific prompt templates used in these baselines are presented in Table  \ref{tab:tabel3}.

\subsection{Main Result}
The main result of different proposed methods is depicted in Tab. \ref{tab:tabel4}. According to the results, the presented MTPrompt show better performance on different types of task in the following aspects: \\
(a) Compared with the fine-tuning and GPT-3 in-context learning method, our MTPrompt could get better performance on different tasks, because our MTPrompt adopts more instructive prompt-based fine-tuning methods, allowing the PLM to learn the minor change on unique tokens. \\
(b) Comparing with LMBFF (discrete prompt), the Meta-prompt could improve the performance of different text classification tasks. Current prompt-based fine-tuning methods need task-related knowledge to motivate the PLM to finish the semantic understanding tasks. \\ 
(c) The performance of the MTPrompt indicates that our proposed prompt could reach more gains from the meta-description, and it also shows that the MTPrompt can find the suitable prompt and offer more augmented prompt choices for prompt learning methods.\\
(d) The effectiveness of SST-2 tasks and QNLI tasks shows the insensitive to binary emotional classification task SST-2 with high accuracy, whereas it is still effective for binary entailment classification with low accuracy. We consider that there is no need to add a particular augmented prompt for high-accuracy tasks or simple tasks since the specific MTPrompt may bring much distracting information.\\
In general, our results show that the proposed MTPrompt could get better improvement compared with different baselines.

\subsection{Explore Different Choices of MTPrompt}

\subsubsection{MTPrompt for Different tasks}
A compact MTPrompt consists of an Object Description (OD), Summary Description (SD), and Task Description (TD). In evaluating MTPrompt, all the descriptions in the MTPrompt used in the proposed paper are listed in Tab. \ref{Tab:C1}. Note that the label mappings in Meta-prompt still use the setting of Tab. \ref{tab:tabel3}.

\subsubsection{Ablation experiments}
To explore different types of task descriptions in Meta-prompt, we further develop ablation experiments on different types of meta-prompt in SST-5 emotion classification tasks. In detail, we use different experimental combinations constructed by three kinds of meta-prompt, object description (OD), summary description (SD), and task description (TD).

As the Tab. \ref{tab:tabel5} depicted, it reviews that these three description types can motivate the PLMs to generate more suitable predictions. With the increase in the description type, the accuracy and median value of the current task could be improved, and its variance is decreased. It indicates that using these three types of description and their combination to construct a prompt can help the PLMs find the task-related knowledge embedded in PLMs and increase the effectiveness of the current task. 

All the meta-description have certificated that PLM can realize the semantic understanding to some degree. We consider that the task-related tokens could optima the representation of prompt projector, which is utilized to map the representation of [mask] tokens to all PLM's vocabulary tokens. Compared with the original probability distribution of mapping tokens, our proposed MTPrompt, which contains more task-driven information, could empower the representation of unique labels, reducing the probability of irrelevant label mapping tokens and useless mapping tokens. 

However, though the median accuracy is increasing, the accuracy begins to reduce, and its variance increases when we combine these three description types. Therefore, using more types of descriptions may sometimes bring more interference when the length of augmented tokens is much longer than the original sentence. It indicates that we should balance the prompt size and the initial input, and some representing methods of these task descriptions could also be considered.

\subsection{Explore the Stability of MTPrompt}
Since relevant studies have proved that different batch sizes and random seeds are the keys to small sample learning experiments, we further evaluated the transferability and stability of the proposed method under different experimental settings. Two sentiment analysis tasks, SST-2 and SST-5, were selected to verify the effects of all experimental Settings, including three different batch sizes and five different random seeds. We use the variable control method, which only changes one experimental setting at a time. 

As shown in Figure \ref{tab:tabel6}, it is revealed that all the proposed methods can exceed the main experimental setting without any large-variance results. Specifically, in the SST-2 experiment, we achieved the best effect at bs=8 and seed=21. In contrast, in sst-5, we achieved the best results at bs=16 and seed=42, which far exceeded the main experiment (mean) effect and the comparison model LMBFF. Our method has certain stability and portability in different experimental Settings and the same task.
\section{Related Works}
\subsection{Prompts in PLMs}
The effectiveness of two-step PLMs mostly depend on its mutual motivation between huge open-domain datasets and small task-oriented data. On the one hand, huge pre-training corpus offer a data foundation and interface for recording and searching external knowledge by the memory parameter. On the other hand, task-oriented data could retain or restrain adjusted parameters to motivate local similar knowledge. Compared to traditional training paradigm, PLMs have achieved better performance in different NLP task and the explanation of PLMs thus become a hot question. One consensus is PLMs could learn the language structure and knowledge existing in large-scale corpus\cite{petroni-2019}. At the same time, the problem of what knowledge and information PLMs learned also attract many attentions\cite{jiang2020can}. Lin et al. found that BERT encodes the location information about the word markers well at its lower level, but switches to the hierarchy-oriented encoding at the higher level\cite{lin2019open}. Jiang et al. propose mining and paraphrase based approaches to automatically generate high-quality and diverse prompts to estimate the knowledge contained in LM. While learning language knowledge, these PLMs models may also store relational knowledge that exists in training data. In addition, the performance of PLMs was largely due to biased hints about which data sets had been fitted, and the prediction is improved mainly through entity guidance and golden answer leakage \cite{cao2021}. Therefore, with the exponential growth in PLMs parameter, PLMs now could seem as a linguistic tool, which could directly generate the prediction of different NLP tasks\cite{radford2019language}.

\subsection{Prompt Fine-tuning}
Roberts et al. found that using natural language queries to fine-tuning the PLMs can also achieve competitive results comparing with extracting the external knowledge\cite{roberts2020}. What is more, given with task description or few in-context data, GPT-3 could efficiently finish many NLP tasks. These shifts current fine-tuning paradigm to prompt fine-tuning, a learning paradigm used different prompt strategy to activate the knowledge generation.
In task-oriented prompt application, Schick\cite{schick2021exploiting,Schick2020} introduces a PET model, which adds a cloze-quesiton template in orgin input, and then predict the masked tag for task of few-shot text classification. Han et al. design the basic sub-prompts manually for text classification and apply logical rules to combine these sub-prompts into the final task-specific prompts\cite{han2021ptr}. Ding et al construct an entity-oriented verbalizer and templates used for fine-grained entity tagging\cite{ding2021entity}. Madotto et al Use only a few examples of each skill and prompt learning to created a end-to-end Bot for question answering \cite{Madotto202110}, and Zhong et al. put forward a OptiPrompt model for factual Probing\cite{zhong2021factual}. In general, these works have certified the power of prompt fine-tuning, and the design of prompt and tuning strategy are significant for each task.
\section{Conclusion}
To alleviate the problem of finding suitable prompts for large language models (like Chat-GPT and GPT-3), we proposed a more instructive prompt learning model -- MTPrompt. Based on the proposed prompt mechanism in language models, we introduce three types of task's meta description to help language motivate the related knowledge(in detail, related tokens). We also put forward two strategies to filter the prompt candidate set. Our results show that MTPrompt could require better improvement on emotion classification, question classification, and natural language inference tasks, which reveals that the proposed MTPrompt could be assumed to be a better knowledge probe of PLMs. Our contrast and supplement experiments also show that MTPrompt has certain stability and portability in different experimental Settings and the same task.

However, the MTPrompt model still has some limitations: (a) The proposed prompt designs are based on experiential tests on different task-oriented metadata descriptions; (b) Though prompt-based fine-tuning method could achieve fast fine-tuning, an automotive and fast searching these tokens should be considered. Our further work would concentrate more on this and offer more technical guidance of MTPrompt.

\bibliographystyle{IEEE}
\bibliography{tprompt}

\end{document}